\pdfoutput=1
\documentclass{article}
\usepackage{hyperref}
\usepackage[dvipsnames, table]{xcolor}  
\usepackage[accepted]{icml}

\usepackage{times}
\usepackage{latexsym}

\usepackage[utf8]{inputenc} % allow utf-8 input
\usepackage[T1]{fontenc}    % use 8-bit T1 fonts
\usepackage{hyperref}       % hyperlinks
\usepackage{url}            % simple URL typesetting
\usepackage{booktabs}       % professional-quality tables
\usepackage{amsfonts}       % blackboard math symbols
\usepackage{nicefrac}       % compact symbols for 1/2, etc.
\usepackage{microtype}      % microtypography
\usepackage{amssymb}
\usepackage{epsfig,graphicx}
\usepackage{multirow,booktabs, hhline}
\usepackage{amsmath, bm}
\usepackage{wrapfig}
\usepackage{lipsum}
\usepackage{xspace}
\usepackage{xcolor}
\usepackage{footnote}
\usepackage[bottom]{footmisc}
\usepackage{microtype}
\usepackage{graphicx}
\usepackage{array}
\usepackage{arydshln}
\usepackage{multirow}

\usepackage{bm}
\usepackage{amssymb}
\usepackage{amsmath}
\usepackage{graphicx}
\usepackage{makecell}
\usepackage{wrapfig}
\usepackage{lipsum}  % generates filler text
\usepackage{etoolbox}
\usepackage{comment}
\usepackage{enumitem}

\usepackage{booktabs}
  
\usepackage{microtype}
\usepackage{csquotes}
\usepackage{bm}
\usepackage{amsmath,amssymb, amsfonts,mathtools}
\usepackage{soul}
\usepackage{adjustbox}
\usepackage[]{todonotes}
\usepackage[capitalize]{cleveref}
\usepackage{hyperref}
\usepackage{svg}
\usepackage{caption}
\usepackage{subcaption}

\newcommand{\method}{\textsc{InstructCTG}\xspace}

\newcommand{\fluency}{\textsc{fluency}}
\newcommand{\success}{\textsc{succ}}

\begin{document}

\icmltitlerunning{Controlled Text Generation with Natural Language Instructions}

\twocolumn[

\icmltitle{Controlled Text Generation with Natural Language Instructions}

\icmlsetsymbol{equal}{*}
                \begin{icmlauthorlist}
                \icmlauthor{Wangchunshu Zhou}{eth}
                \icmlauthor{Yuchen Eleanor Jiang}{eth}
                \icmlauthor{Ethan Wilcox}{eth}
                \icmlauthor{Ryan Cotterell}{eth}
                \icmlauthor{Mrinmaya Sachan}{eth}
                \end{icmlauthorlist}
                
                \icmlaffiliation{eth}{ETH Z{\"u}rich}
                %\icmlaffiliation{comp}{Company Name, Location, Country}
                \icmlcorrespondingauthor{Wangchunshu Zhou}{wangchunshu.zhou@inf.ethz.ch}

% You may provide any keywords that you
% find helpful for describing your paper; these are used to populate
% the "keywords" metadata in the PDF but will not be shown in the document
\icmlkeywords{Machine Learning, ICML}

\vskip 0.3in
]
\printAffiliationsAndNotice{}

\begin{abstract}
Large language models can be prompted to produce fluent output for a wide range of tasks \emph{without} being specifically trained to do so.
Nevertheless, it is notoriously difficult to control their generation in such a way that it satisfies user-specified constraints. 
In this paper, we present \method, a simple controlled text generation framework that incorporates different constraints by verbalizing them as natural language instructions. 
We annotate natural texts through a combination of off-the-shelf NLP tools and simple heuristics with the linguistic and extra-linguistic constraints they  satisfy. 
Then, we verbalize the constraints into natural language instructions to form weakly supervised training data, i.e., we prepend the natural language verbalizations of the constraints in front of their corresponding natural language sentences.
Next, we fine-tune a pre-trained language model on the augmented corpus.
Compared to existing methods, \method is more flexible in terms of the types of constraints it allows the practitioner to use.
It also does not require any modification of the decoding procedure.
Finally, \method allows the model to adapt to new constraints without re-training through the use of in-context learning.
Our code is available at \url{https://github.com/MichaelZhouwang/InstructCTG}.

\end{abstract}

\section{Introduction}
Large language models (LLMs) pre-trained on web-scale corpora~\citep{gpt,gpt2,gpt3} have demonstrated the ability to generate fluent and realistic text.
Yet, many text generation applications require the model to generate text that not only possesses a high degree of naturalness but also adheres to task-specific constraints.
For instance, a question generation system may require that the model incorporates certain keywords during generation, and a dialog system may be expected to generate responses that have a certain tone.\looseness=-1

\begin{figure*}%[t]
   \centering
    \vspace{-5pt}
  \includegraphics[width=0.85\textwidth,trim={0cm 0.1cm 0cm 0.8cm} ,clip]{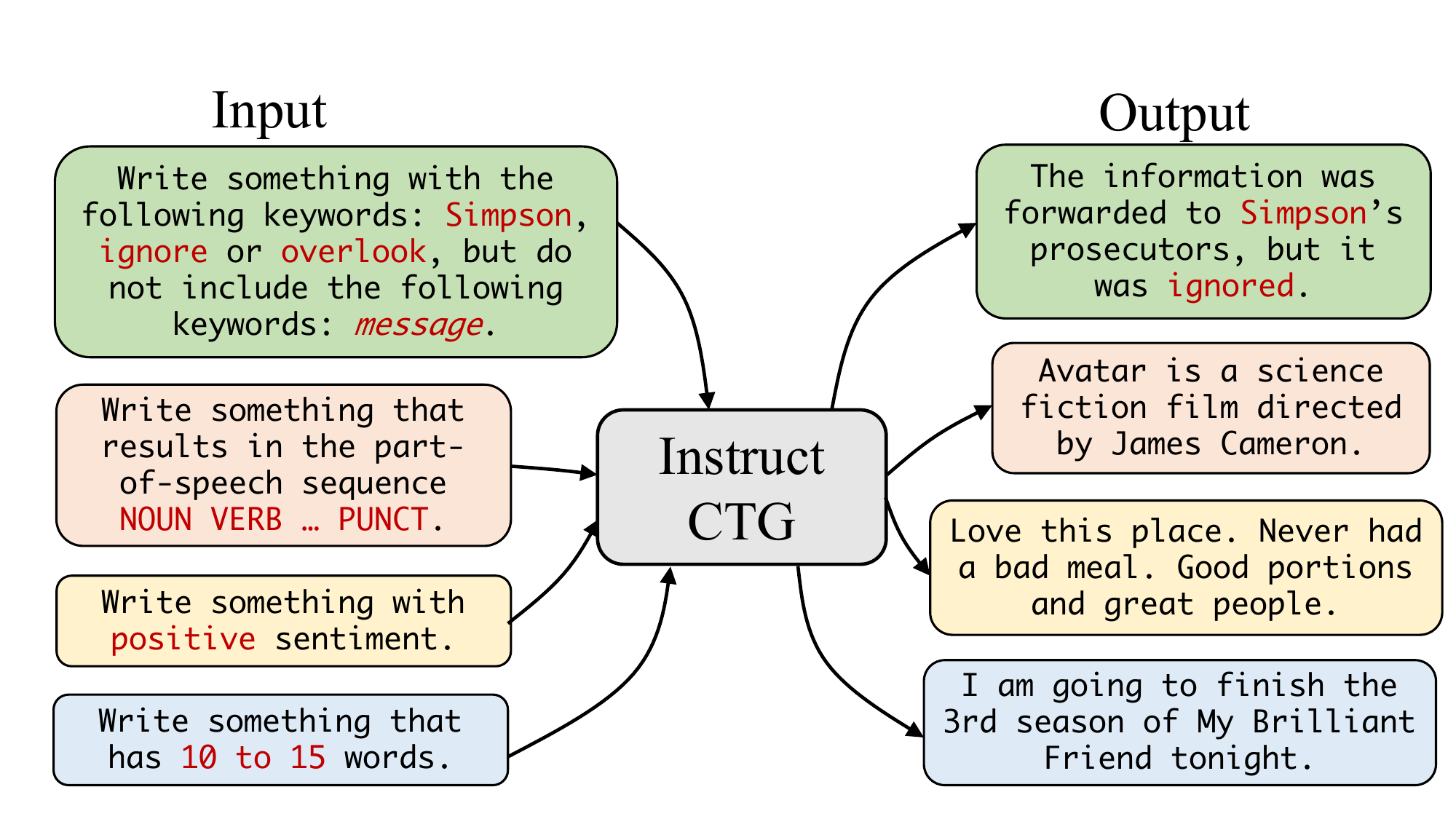}
    \vspace{-6pt}
   \caption{A cartoon of the \method framework. \method is an instruction-tuned model for controlled text generation. In our experiments, we consider \textsc{lexical}, \textsc{syntax}, \textsc{semantic}, \textsc{style}, and \textsc{length} constraints.\looseness=-1}
  \vspace{-10pt}
   \label{fig:motivation}
\end{figure*}

The dominant paradigm for controlled text generation involves incorporating the constraints into the decoding algorithm itself.
Prior work either includes constraints in the search algorithm by modifying beam search~\citep{hokamp-liu-2017-lexically, anderson-etal-2017-guided,post-vilar-2018-fast,lu-etal-2021-neurologic2,lu-etal-2022-neurologic2} or by converting the constraints into classifiers, which are then used to guide the decoding process \citep{qin2022cold,kumar2022gradient,li2022diffusionlm,amini2023structured}. 
While effective at requiring the model to enforce the constraints on the generated text, existing decoding-time methods for controlled text generation suffer from a number of limitations. 
First, decoding-time methods modify the output sequence distribution and, thus, can result in lower-quality text~\citep{lin-etal-2020-commongen}.
Second, they may require extra computation and lead to slower generation speed~\citep{post-vilar-2018-fast,li2022diffusionlm}.
Third, they are often less able to deal with new kinds of constraints or a novel composition of existing constraints. 
Finally, they may require practitioners to design new, bespoke features or train additional classifiers to accommodate novel constraints.\looseness=-1

In this paper, we present \method, a simple paradigm for controlled generation, which presents an alternative to decoding-time-constrained generation methods.  
In contrast, we adopt a training-time approach.
Under \method, we construct a corpus in a weakly supervised fashion where each sentence in the corpus is treated as if it had been constructed under the influence of one or more latent constraints.
We automatically impute these latent constraints using off-the-shelf NLP tools or simple rules.
Then, \method incorporates the constraints into generation through a prompt that verbalizes the constraint into a natural language instruction.
As a concrete example, in order to generate sentences with a length constraint, we collect a large corpus of sentences and automatically derive the length of each sentence by using a pre-trained tokenizer and counting the number of tokens.
Then, in order to conduct controlled generation, we verbalize the length constraints as natural language instructions and instruction-tune the model using the augmented corpus~\citep{wei2022finetuned,sanh2022multitask}.
To avoid the cost of always having to fine-tune for every constraint, we also use meta in-context learning~\citep{min-etal-2022-metaicl} for some constraints where we only make use of a few demonstrations of the constraint--text pairs.
Compared to methods that incorporate constraints into the decoding algorithm directly, \method gives the model access to the constraints before decoding, leading to improved generation quality and efficiency.
A cartoon illustration of \method is given in \cref{fig:motivation}.\looseness=-1

To evaluate the effectiveness of \method, we conduct a wide range of experiments with different constraint types including keywords, syntax, semantic, style, and length, by fine-tuning T0~\citep{sanh2022multitask}, an instruction-tuned text-to-text pre-trained model.
Our results show that \method achieves a high level of constraint satisfaction that is comparable to or better than existing decoding-based methods.
The method is also faster in some cases.
Additionally, \method demonstrates strong generalization abilities in the context of few-shot learning, specifically with regard to the constraints that are unseen during training.\looseness=-1

\section{\method}
Our proposed framework, \textit{Instruction-based Controlled Text Generation} (\method), is a prompt-based controlled text generation framework.
\method verbalizes constraints into natural language and uses instructions that consist of the natural language verbalization and a few demonstrations to encourage the model to incorporate the constraints during generation. 
In this section, we describe the data synthesis method, input format, training method, and inference method of \method.\looseness=-1

\subsection{Data Collection}
\label{sec:data}
Typically, instruction tuning requires a large amount of labeled training data.
However, most publicly available text generation datasets do not have constraint annotations.
And, moreover, crowdsourcing a sufficient number of constraints--text pairs for instruction tuning would be expensive and time-consuming.
With this as the state of play, we propose a weakly supervised data synthesis method to synthesize constraint--sentence pairs for \method.
The main idea behind our data synthesis strategy is to view various kinds of constraints as latent prompts that implicitly inspired the generation of the naturally occurring sentences.
However, instead of attempting to infer the latent prompts through, say, a sophisticated Bayesian model, we opt instead to use simple rules and off-the-shelf NLP tools to impute some approximation to the latent prompts. 
We next describe the constraints used in the training stage of \method.\looseness=-1

\paragraph{a) Lexical Constraints} Following previous work on constrained decoding, we refer to those constraints that require the generated text to include certain keywords in the output text as lexical constraints.
To construct lexical constraints, we adopt the keyword extraction pipeline used in prior work on keyword-based text generation~\citep{lin-etal-2020-commongen,zhou2021pretraining}. Specifically, given an input text sequence, we use a part-of-speech tagger\footnote{{\tiny \url{https://spacy.io/models/en\#en\_core\_web\_sm}.}} to extract tokens with the tags that encode some form of a noun or verb from the sentence and randomly sample a portion of them as keywords. 
We further lemmatize the extracted keywords for consistency across morphological variants.
We also design a data augmentation method to enable \method to handle more complex requirements such as choosing one keyword from a set of keywords or not including a certain word in the output. 
Specifically, for each keyword, we use a BERT-based lexical substitution system~\citep{zhou-etal-2019-bert} to generate possible alternatives of the keyword in the context. 
We then create two more complex types of lexical constraints.
The first is a disjunctive constraint that requires that the text must include (at least) one of the keywords from the set of the original and the alternative words.
The second is an exclusion constraint that requires the text to \emph{not} include any of the proposed alternative keywords.
We randomly sample sentences from the C4~\citep{raffel2020exploring} dataset to synthesize our keyword constraint--text pairs.\looseness=-1

\paragraph{b) Syntactic Constraints}
We also consider syntactic constraints that require the output text to conform to certain syntactic dictates.
Specifically, we focus on two types of syntactic constraints: \textbf{part-of-speech} sequence constraints and \textbf{syntactic tree} constraints. 
Our part-of-speech (POS) constraints are   sequences of POS tags (e.g., \textsc{poss} \textsc{verb} \textsc{det}
\textsc{noun}) that require the model to generate a sequence of words of the same length whose POS tags match the target (e.g., ``He bought a car''). 
Similarly, syntactic tree constraints consist of a linearized syntactic parse tree (e.g., (S (NP (PRP $*$)) (VP (VBD $*$) (NP (DT $*$) (NN $*$)))) where $*$ denotes a wild card) and require the model to generate a text sequence that matches the constraint when parsed with a pre-trained parser~\citep{kitaev-klein-2018-constituency}. 
Similar to the generation of the lexical constraints, we randomly sample sentences from the C4 dataset to synthesize syntactic constraint--text pairs.

\paragraph{c) Semantic Constraints}
To enable fine-grained control over the content and semantics of the output text, we also include three semantic constraints including a {\bf topic/domain constraint}, a {\bf sentiment constraint}, and a {\bf political slant constraint}. 
Concretely, the topic constraint requires the model to produce texts related to a certain topic, for example, a text about a certain scientific field. 
Similarly, the sentiment constraint and the political slant constraint control the sentiment polarity and the political slant of the output texts, respectively.
We first train a RoBERTa-based classifier \citep{liu2019roberta} on labeled datasets for different semantic constraints with an additional \textsc{neutral} output class which is associated with random sentences from the C4 dataset.
For instance, to enforce a sentiment constraint, we train a three-way classifier on sentences marked with \textsc{positive}, \textsc{negative}, or \textsc{neutral}.
The sentences labeled with \textsc{positive} and \textsc{negative} are taken from the Amazon Review dataset~\citep{he2016ups} and
the sentences labeled with \textsc{neutral} class are drawn randomly from the C4 dataset and additionally filtered by a two-way classifier trained on the Amazon Review dataset to have no sentiment polarity.
To simulate the population of sentiment-polarized sentences in web text, we construct the train set with 90\% random sentences and 5\% sentences of positive and negative sentiment each.
We then apply the trained classifier on more randomly sampled sentences from C4 and use sentences that are labeled to be of positive or negative sentiment with high confidence to synthesize sentiment constraint--text pairs. 
Additionally, we use the Political Slant dataset~\citep{voigt2018rtgender} for the political slant constraint and use the M2D2 dataset~\citep{reid2022m2d2}, which contains diverse topics in Wikipedia and arXiv categories, for topic control.

\paragraph{d) Style Constraints}
Apart from controlling keywords, syntax, and content, it is also desirable to control the style of the generated text.
Here, we adopt the categories outlined in \citet{jin-etal-2022-deep} that include (\textit{inter alia}) the formality, politeness, biasedness, toxicity, and simplicity of the output texts. 
Similar to sentiment constraints, we also train a style classifier with a \textsc{noStyle} class with random sentences and collect stylized sentences from the C4 dataset to construct style constraint--text pairs. 
We use Grammarly’s Yahoo Answers Formality Corpus (GYAFC) dataset~\citep{rao-tetreault-2018-dear} for formality constraints, the politeness transfer dataset collected by \citet{madaan-etal-2020-politeness} for politeness control, the FlickrStyle stylized caption
dataset~\citep{gan2017stylenet,li-etal-2018-delete} for the control of style, the Wiki Neutrality Corpus~\citep{pryzant2020automatically} for biasedness control, the PWKP~\citep{zhu-etal-2010-monolingual2} dataset for text simplification, and following the data collection method in \citet{tran-etal-2020-towards} for collecting offensive/non-offensive sentences from the C4 dataset.

\paragraph{e) Length Constraints}
Finally, we also consider a constraint that controls the length of the output text. The length constraint is simply the number of words of the output text after tokenization (measured using the Moses~\citep{koehn-etal-2007-moses} tokenizer). 
We consider an interval-based length constraint with an interval of $5$, which requires the output text to contain between $5n$ and $5(n+1)$ words for a given integer $n$.
We construct length constraint--text pairs using randomly sampled sentences from C4 and the Moses tokenizer.

Using the process described above, we synthesize 1 million training constraints--text pairs for each category of constraint (lexical, syntactic, semantic, style, and length) and 50,000 pairs, each, for the development and test sets.\looseness=-1

\newcommand{\instruct}[1]{\textcolor{BrickRed}{#1}}
\newcommand{\concolor}[1]{\textcolor{gray}{#1}}
\newcommand{\STAB}[1]{\begin{tabular}{@{}c@{}}#1\end{tabular}}
\newcommand{\task}[1]{\multirow{4}{*}{\STAB{\rotatebox[origin=c]{90}{\textsc{#1}}}}}

\begin{table*}[t]
\centering
\begin{adjustbox}{width=\textwidth}
\begin{tabular}{ccp{440pt}}
\toprule[2pt]
% \multicolumn{2}{c}{\textsc{lexical}} \\
% \midrule[1pt]
\task{lexical} & \cellcolor{ForestGreen!20} \textsc{target}  & \cellcolor{ForestGreen!20}The information was forwarded to Simpson's prosecutors, but it was ignored. \\
& \cellcolor{gray!20} \textsc{constraints}  & \cellcolor{gray!20} Simpson   $\wedge$ (ignore $\vee$ overlook) $\wedge$ information $\wedge$ prosecutor $\vee$ $\neg$ message \\
& \textsc{instruction-1} & \instruct{Write something with the following keywords:} Simpson, ignore \instruct{or} overlook, information, prosecutor, \instruct{but do not include the following keywords:} message.  \\
& \textsc{instruction-2} & \instruct{Write something that contains the following words:} Simpson, ignore \instruct{or} overlook, information, prosecutor \instruct{but, does not contain the word} message.  \\
\midrule[2pt]
% \multicolumn{2}{c}{\textsc{syntax}} \\
% \midrule[1pt]
\task{syntax} & \cellcolor{ForestGreen!20} \textsc{target}  & \cellcolor{ForestGreen!20}Avatar is a science fiction film directed by James Cameron. \\
& \cellcolor{gray!20} \textsc{constraints}  & \cellcolor{gray!20}  \textsc{noun} \textsc{verb} \textsc{det} \textsc{noun} \textsc{noun} \textsc{noun} \textsc{verb} \textsc{prep} \textsc{noun} \textsc{noun} \textsc{punct} \\
& \textsc{instruction-1} & \instruct{Write something with a part-of-speech sequence} \textsc{noun} \textsc{verb} \textsc{det} \textsc{noun} \textsc{noun} \textsc{noun} \textsc{verb} \textsc{prep} \textsc{noun} \textsc{noun} \textsc{punct}\instruct{.}  \\
& \textsc{instruction-2} & \instruct{Write something that results in the part-of-speech sequence} \textsc{noun} \textsc{verb} \textsc{det} \textsc{noun} \textsc{noun} \textsc{noun} \textsc{verb} \textsc{prep} \textsc{noun} \textsc{noun} \textsc{punct} \instruct{after part-of-speech tagging with Spacy.} \\
\midrule[2pt]
% \multicolumn{2}{c}{\textsc{semantic}} \\
% \midrule[1pt]
\task{semantic} & \cellcolor{ForestGreen!20} \textsc{target}  & \cellcolor{ForestGreen!20}Love this place. Never had a bad meal. Good portions and great people.\\
& \cellcolor{gray!20} \textsc{constraints}  & \cellcolor{gray!20}  the sentiment is positive \\
& \textsc{instruction-1} & \instruct{Write something with} positive \instruct{sentiment}\instruct{.}  \\
& \textsc{instruction-2} & \instruct{Write something} positive\instruct{.} \\
\midrule[2pt]
% \multicolumn{2}{c}{\textsc{sytle}} \\
% \midrule[1pt]
\task{style} & \cellcolor{ForestGreen!20} \textsc{target}  &  \cellcolor{ForestGreen!20}I dunno, but gonna give a try.\\
&\cellcolor{gray!20} \textsc{constraints}  & \cellcolor{gray!20} the style is informal\\
&\textsc{instruction-1} &\instruct{Write something that is} informal\instruct{.}  \\
&\textsc{instruction-2} & \instruct{Write something in an} informal \instruct{tone.}\\
\midrule[2pt]
% \multicolumn{2}{c}{\textsc{length}} \\
% \midrule[1pt]
\task{length} & \cellcolor{ForestGreen!20} \textsc{target}  &  \cellcolor{ForestGreen!20}I am going to finish the 3rd season of My Brilliant Friend tonight. \\
& \cellcolor{gray!20} \textsc{constraints}  & \cellcolor{gray!20} length of 10-15 words \\
& \textsc{instruction-1} &\instruct{Write something that has} 10 to 15 \instruct{words.}  \\
& \textsc{instruction-2} &\instruct{Write something with} 10 to 15 \instruct{words}\\
\bottomrule[2pt]
\end{tabular}
\end{adjustbox}
    \caption{Example texts, constraints, and the corresponding verbalized instructions. 
    The instructions are highlighted in red. Note that there are multiple possible ways to verbalize instructions for each task.\looseness=-1}
    \label{tab:examples}
   % \vspace{-15pt}
\end{table*}

\subsection{Constraint Verbalization}
\label{sec:verbalization}

We verbalize the automatically derived constraints in different natural language formats.
As shown in Table \ref{tab:examples}, for each constraint type, we design natural language templates to verbalize the constraints into natural language prompts. 
Verbalizing constraints has a few benefits: First, verbalizing the constraints in natural language effectively exploits the prompt-based generation ability of the pre-trained model.
Second, instruction-tuning with natural language verbalization of the constraints enables zero-shot constraint generalization, enabling the model to handle new constraints that are unseen during training by simply describing them in natural language. Moreover, we design multiple diverse natural language templates for each constraint used for training. 

\subsection{Composition of Constraints}
\label{sec:composition}
Controlled text generation applications often require more than one constraint to be enforced at the same time during generation.
To enable \method to handle such a composition of constraints, we verbalize the constraints individually and conjoin them with the conjunctive ``and''.
For example, for the sentence ``He bought a car'' with both lexical and POS constraints, the final instruction will be ``write something with keywords \textit{buy}, \textit{car} that has a part-of-speech tag sequence \textsc{prp} \textsc{verb} \textsc{det} \textsc{noun}.''
We subsample the C4 dataset again to obtain another 1 million training examples with multiple constraints.
For each sampled text, we first try to produce constraint verbalizations for as many constraint types as possible and then randomly sample a subset of the verbalizations with between 2 to 5 constraints uniformly at random.\looseness=-1

\subsection{Extension to Conditional Text Generation}
\label{sec:conditional}
The aforementioned data synthesis pipeline as described is applicable to unconditioned text generation. 
However, several major applications of text generation (e.g., paraphrase generation, summarization, etc.) deal with the generation of text conditioned on another piece of text. Therefore, we also extend our controlled generation pipeline to conditional text generation tasks.

In order to achieve this, instead of randomly sampling texts from a general unlabeled corpus like C4, we instead turn to datasets used for various conditional text generation tasks like paraphrase generation and summarization.
In order to generate instructions for controllable conditional generation, we
compose the original task-specific prompt with our natural language instructions for the set of control constraints.
For instance, consider the following training instance for paraphrase generation: ``How far is the earth from the sun?'' $\mapsto$ ``What is the distance between the sun and the earth?''.
In this case, the augmented instruction for a lexical constraint would be: ``Write a paraphrase of `how far is the earth from the sun with keyword \textit{distance}' ''.
Our experiments consider paraphrase generation and question generation tasks and use the Quora Question Paraphrase dataset and the SQUAD question generation dataset~\citep{rajpurkar-etal-2016-squad,du-cardie-2018-harvesting}, respectively.\looseness=-1

\subsection{Instruction-based Meta In-context Learning}
\label{sec:metaicl}
After the above data synthesis process, we use a combination of instruction tuning~\citep{wei2022finetuned} and meta in-context learning~\citep{min-etal-2022-metaicl} where we prepend the natural language instructions for the constraints with 5 demonstrations, i.e., constraint--output pairs, of the same constraint or a composition of several constraints.
To enable the model to do controlled text generation both with and without demonstrations, we omit the demonstrations from 50\% of the examples during training. 
In those cases, we just perform instruction tuning.
Given the collection of instructions (and demonstrations), we fine-tune the language model using maximum likelihood estimation and teacher forcing.\looseness=-1

\subsection{Decoding}
At test time, unlike most previous work that requires complex decoding algorithms, \method can perform controlled text generation without any modification to the decoding process. Specifically, we can simply verbalize the target constraints (either using existing templates for seen constraints or writing a new description for unseen constraints) and optionally provide a few examples for the target constraint types. 
Then we can simply apply an out-of-box generation method, e.g., beam search.\looseness=-1

\section{Experiments}
We now turn to the empirical portion of our paper.\looseness=-1
\subsection{Experimental Setup}

\begin{table*}
\centering
\resizebox{1.0\textwidth}{!}{
\begin{tabular}{lcc|cc|cc|cc|cc|c}
\toprule
          & \multicolumn{2}{c|}{\textsc{lexical}} & \multicolumn{2}{c|}{\textsc{syntax}} & \multicolumn{2}{c|}{\textsc{semantic}} & \multicolumn{2}{c|}{\textsc{style}} & \multicolumn{2}{c|}{\textsc{length}} & 
          \multicolumn{1}{c}{\textsc{inference}} \\
& \success        & \fluency $\downarrow$       & \success     & \fluency $\downarrow$   & \success     & \fluency $\downarrow$    & \success     & \fluency $\downarrow$   &  \success     & \fluency $\downarrow$ & \textsc{time}  \\ \midrule
CFT  & 95.1  & 16.7  & 85.7 & 17.9 & 85.1 & 15.2 & 87.6 & 17.4  & \bf 100.0 & 15.9 & 1.0 $\times$\\
DBA  & 96.8   & 23.4  &  - & - & - & - & - & - & - &  - & 1.3 $\times$ \\
NeuroLogic  & \bf 97.9   & 21.8  &  - & - & - & - & - & - & - & - & 4.8 $\times$ \\
COLD decoding     & 95.1 & 41.3  & 82.2 & 48.5  & 80.9   & 38.9 & 82.3 & 43.0 & -  & - & 34.8 $\times$  \\
Diffusion-LM & 94.3 & 33.2  & \bf 88.6 & 35.9  & 82.8  & 31.4 & 85.9 & 32.6 & 99.8  & 27.7 & 48.3 $\times$  \\
\midrule
\method  & 97.5  & \bf 12.6  & 88.3 & \bf 14.4 & \bf 88.2  & \bf 12.2 & \bf 90.6 & \bf 13.1  & \bf 100.0 & \bf 11.9 & 1.1 $\times$ \\
\ w/o verbalization  & 95.5  & 16.3  & 85.0 & 17.9    & 85.5   & 14.7 & 88.4 & 15.9  & \bf 95.3 & 14.1 & 1.1 $\times$ \\
\ w/o demonstrations & 96.8  & 13.3  & 87.8 & 15.3 & 87.8   & 13.0 & 90.1 & 13.6  & \bf 100.0 & 12.2 & 1.0 $\times$ \\
\ w/o multi-constraint training  & 97.2  & 12.9  & 87.0 & 14.9 & 86.8  & 13.1 & 89.2 & 14.2  & \bf 100.0 & 12.4 & 1.1 $\times$  \\
\ w/o multiple templates & 97.3  & 12.8  & 88.1 & 14.6    & 84.7   & 12.8 & 90.2 & 13.9  & \bf 100.0 & \bf 11.9 & 1.1 $\times$ \\
\bottomrule
\end{tabular}}
\caption{\label{tab:results_seen} Results on constraints seen during training. \method is faster and achieves better fluency (\fluency) compared to all baselines while maintaining competitive constraint satisfaction rate (\success) across all 5 constraint types.}
\vspace{-5pt}
\end{table*}

\paragraph{Model} We use T0-11B as our base model because it has previously been instruction-tuned on many NLP tasks.
T0-11B is initialized from the 11B parameter version of T5+LM~\citep{raffel2020exploring}, which is pre-trained on the C4 dataset with a text infilling objective and instruction-tuned on a collection of various NLP datasets.\looseness=-1

\paragraph{Training}
\label{sec:training}
Our model is trained on the mixture of training constraint types illustrated in Table \ref{tab:examples}. 
We hold out one semantic constraint (political slant) and two style constraints (humor and politeness) to evaluate \method's ability to generalize to unseen constraints.
We fine-tune on all other constraints. 
Following \citet{raffel2020exploring}, we assemble our multi-constraints training mixture by combining and shuffling all examples from all training constraints. 
We fine-tune our model with the Adam~\citep{kingma2014adam} optimizer for 100000 steps with a learning rate of 1e-4, a batch size of 1024 text pairs, a dropout rate of 0.1, and a learning rate warmup of 8000 steps.
Following \citet{sanh2022multitask}, we perform checkpoint selection by choosing the checkpoint with the highest constraint satisfaction rate (see the next paragraph) on the validation splits of datasets of training constraints.\looseness=-1

\paragraph{Evaluation}
We consider two evaluation metrics: constraint satisfaction rate and fluency.
The constraint satisfaction rate is computed differently for each type of constraint.
For the lexical and length constraints, we take it to be exact match, for the part-of-speech sequence constraint, we take it to be token-level accuracy, and, for the syntactic tree constraint, we take it to be labeled constituent $F_1$, and, finally, for the semantic and style constraints, we train a RoBERTa-based classifier to evaluate the controlled generation, following previous work on text style transfer~\citep{xu-etal-2018-unpaired}.
For our fluency metric, we use the perplexity of the generated text under OPT-30B, another language model trained on web text~\citep{zhang2022opt}.\looseness=-1

\paragraph{Baselines}
We compare \method with both constrained search algorithms and score-based controlled generative models which are detailed below.
\begin{itemize}
\item \textbf{Dynamic Beam Allocation} \citep[DBA;][]{post-vilar-2018-fast} is a popular constrained search algorithm for lexically constrained text generation. 
It tracks constraint satisfaction status using a finite-state automaton.

\item \textbf{NeuroLogic Decoding} \citep{lu-etal-2021-neurologic2} is one of the state-of-the-art constrained search algorithms. 
NeuroLogic decoding takes logic-based lexical constraints as input and uses prefix tries to track constraint satisfaction.\looseness=-1

\item \textbf{COLD} \citep{qin2022cold} is a scored-based constrained generation method that incorporates various kinds of constraints as energy functions and then performs constraint satisfaction through gradient-based sampling with Langevin dynamics in the logit space.

\item \textbf{Diffusion-LM} \citep{li2022diffusionlm} is a diffusion model that differentiable constraints by considering them as a part of score functions.

\item In \textbf{Constraint-specific fine-tuning (CFT)}, we fine-tune the model on the constraint--text pairs for each constraint type.\looseness=-1
\end{itemize}

To ensure a fair comparison for CBS and NeuroLogic decoding, we fine-tune the pre-trained T0-11B model on the target sentences used for \method training with the unconditional text generation objective because they do not require constraint-specific training. 
For COLD, we use GPT-J as the backbone model because it requires a decoder-only language model for generation. 
For the Diffusion-LM, we initialize the transformer model with T0-11B parameters and train it on the same unlabeled data that is used for \method training. 
We also train a model with the same configuration as described in \citet{li2022diffusionlm} with the same data.
However, we found that it substantially underperforms with respect to our implementation. 
For the CFT baseline, we fine-tune T0-11B on the corresponding datasets for each constraint type.\looseness=-1

\subsection{Results on Seen Constraints}

We first present the performance of \method on seen constraints in Table \ref{tab:results_seen}. 
We observe that \method achieves a comparable or better constraint satisfaction rate compared to many controlled text generation methods while generating much more fluent text.

Additionally, \method is over 4$\times$ faster than NeuroLogic decoding, a strong search-based algorithm, and over 40$\times$ faster than Diffusion-LM.\looseness=-1

\subsection{Results on Unseen Constraints}

We also evaluate the performance of \method with a few demonstrations of constraints that are unseen during training. This is a unique feature of \method since existing search-based algorithms are only capable of lexical constraints and score-based methods require score functions trained on the train set of the target constraint. We present the results in Table \ref{tab:results_unseen}, where we combine the few-shot performance of \method with the supervised results of existing methods. We find that despite only a few demonstrations and no update to the model, \method achieves a higher constraint satisfaction rate and better output quality compared to the baselines.
Moreover, \method also performs competitively with CFT, an oracle baseline that fine-tunes the model on the unseen constraints. This shows that \method has strong few-shot constraint generalization ability, allowing the model to handle new constraints by only writing a description and a few examples of them. 

We also conduct an analysis of the in-domain and the out-of-domain generalization ability of \method. For in-domain tests, we compare \method and the CFT baseline on texts with lengths greater than 60 words; both were trained only on texts whose lengths were less than 60 words. 
We find that CFT’s accuracy drops from 100\% to 91\% when increasing the length bin to 100-105 words while \method’s accuracy only drops to 97\%.

For an out-of-domain test, we consider a variant of \method that is trained on a mixture of lexical, syntactic, content, and style constraints, but not on the length constraint. 
We also consider the original T0-11B model. 
We find that \method results in 82\% of the constraints satisfied without demonstrations, and 88\% of the constraints satisfied with 5 demonstrations.
In contrast, T0-11B only results in 65\% of the constraints satisfied. This confirms that \method helps generalize on out-domain unseen constraints above baselines. These results highlight the importance of demonstrations, which is in line with findings in Table \ref{tab:results_unseen}.\looseness=-1

\begin{table}
\centering
\resizebox{.5\textwidth}{!}{
\begin{tabular}{lcc|cc}
\toprule
          & \multicolumn{2}{c|}{\textsc{semantic}} & \multicolumn{2}{c}{\textsc{style}}   \\
& \success        & \fluency       & \success     & \fluency \\ \midrule
\textcolor{gray}{CFT (Oracle)}   & \textcolor{gray}{84.7}   & \textcolor{gray}{15.8} & \textcolor{gray}{88.1} & \textcolor{gray}{16.6} \\
COLD decoding  & 80.5 & 41.2  & 81.3 & 42.1 \\
Diffusion-LM & 82.1 & 33.9  & 85.3 & 30.3 \\
\midrule
\method  & \bf 83.4  & \bf 15.3  & \bf 87.3 & \bf 14.7   \\
\ w/o verbalization   & 67.7  & 19.8  & 65.1 & 18.5 \\
\ w/o demonstrations  & 79.8  & 16.5  & 83.1 & 15.7 \\
\ w/o multiple templates   & 82.7  & 16.1  & 86.5 & 15.1 \\
\bottomrule
\end{tabular}
}
\caption{\label{tab:results_unseen} Results on unseen constraints. 
We observe that \method achieves better performance compared to the score-based baselines trained on the unseen constraint datasets. Note that, in this setting, CFT is a skyline as it was \emph{trained} constraints unseen by the other methods.}
%\vspace{-15pt}
\end{table}

\subsection{Analysis}

\textbf{Ablation Study}
We first ablate different design choices of \method to investigate their relative importance. The results are shown in Table \ref{tab:results_seen} and \ref{tab:results_unseen}. 
First, we find that training the model without the verbalizing constraints into natural language instructions leads to a substantial performance drop, especially on unseen constraints.
%We believe this is due to the fact that the language model was pre-trained on natural language text.
%training on verbalized constraint instructions encourages the model to better learn the abstract relationship between instruction and output, thus better generalizing to instructions for new constraints. 
Using multiple templates for verbalization also (marginally) improves performance. 
We also find that meta in-context learning helps the model generalize to unseen tasks. 
We hypothesize that this is because examples during training help the model learn content-invariant relationships between examples, constraints, and outputs. 
One can also observe that training with multiple constraints at the same time leads to improved performance. 
In addition, in preliminary experiments, we consider different numbers (1, 2, 5, and 10) of demonstrations and find that 5 demonstrations results in a reasonable performance--efficiency trade-off.\looseness=-1

\paragraph{Results on Conditional Generation Tasks}
We also test \method on conditional natural language generation tasks.
To do so, we further fine-tune our model on two conditional text generation tasks and lexical constraint using \method following the procedure described in section \ref{sec:conditional}. The results are shown in Table \ref{tab:results_conditional}. We can see that \method achieves a comparable constraint satisfaction rate compared to previously proposed controlled generation methods while leading to substantial improvement in task-specific metrics. This confirms the effectiveness of \method on conditional text generation tasks.

\begin{figure}
    \centering
    %\vspace{-2pt}
     % \imgtrimleftsize{} \imgtrimbottomsize{} \imgtrimrightsize{} \imgtrimtopsize
   \includegraphics[width=\columnwidth,trim={0cm 0.1cm 0cm 1.2cm} ,clip]{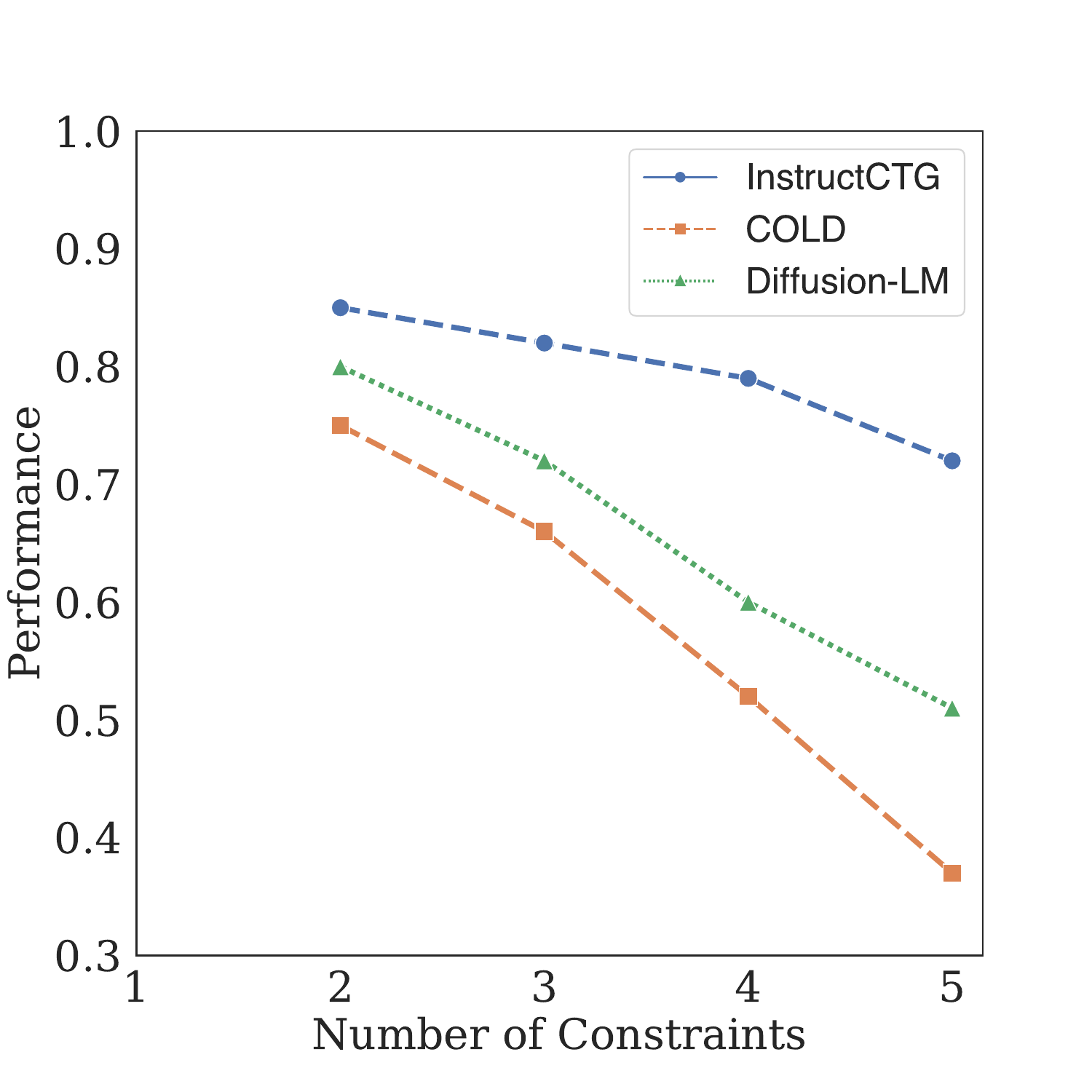}
    \caption{Results on the composition of multiple constraints. 
    The aggregate performance number on the $y$-axis is measured by the following procedure: First, for each constraint type, we calculate the percentage of \success \ and \fluency \ that the model retains compared to the performance without constraint composition. We then average the percentages across all constraint types as the performance number.
    %For score-based baselines, the relative weights are tuned with random search.
    }
    \label{fig:constraint_composition}
\end{figure}

\textbf{Results on Composition of Constraints}
\label{sec:compose_eval}
We also analyze the performance of \method under a composition of constraints. The results are shown in Figure \ref{fig:constraint_composition}. 
%We find that \method is much more robust to a composition of constraints while also alleviating the procedure of searching relative weights. 
The results demonstrate the ability of \method to effectively handle multiple constraints at the same time.\looseness=-1

\section{Related Work}
We now turn to discuss related work.
\subsection{Neural text generation}
Neural text generation generally involves two stages: training and decoding. 
During the training stage, one estimates a probability model over natural language text.
The probability model is typically parameterized autoregressively, i.e., it is factored into the product of per-token conditional probabilities in left-to-right order. 
The per-token conditional probabilities are parameterized with a shared neural network such as an LSTM~\citep{hochreiter1997long} or a Transformer~\citep{transformer}. 
The parameters of the neural network are typically estimated by regularized maximum likelihood estimation.
During the decoding stage, one aims to generate text using the trained model.
Popular decoding algorithms include beam search, top-$k$ sampling~\citep{fan2018hierarchical}, nucleus sampling~\citep{holtzman2019curious}, and locally typical decoding~\citep{meister2022locally}.\looseness=-1
%, which are optimized for the quality--diversity trade-off.
%$\by = y_1 y_2 \cdots y_T$ in the language $\lang \subseteq\kleene{\vocab}$ 

\begin{table}
\centering
\resizebox{.5\textwidth}{!}{
\begin{tabular}{lcc|cc}
\toprule
          & \multicolumn{2}{c|}{\textsc{pharaphrase}} & \multicolumn{2}{c}{\textsc{QG}}   \\
& \success        & \textsc{rouge}      & \success     & \textsc{meteor}
\\ \midrule
FT  & -  & 51.7  & - & 29.4 \\
\ + NeuroLogic decoding     & \bf 96.5 & 48.5  & \bf 95.8  & 28.6  \\
Diffusion-LM & 93.5  & 45.3 & 93.1 & 26.6 \\
CFT  & 94.7  & 55.5  & 94.5 & 33.9 \\
\method  & 96.1  & \bf 57.4  & \bf 95.8 & \bf 35.4   \\
\bottomrule
\end{tabular}
}
\caption{\label{tab:results_conditional} Results on conditional natural language generation tasks. Compared with the baselines, \method achieves better task-specific performance while having comparable accuracy (\success).}
\vspace{-0.4cm}
\end{table}

\subsection{Controlled text generation}
Conventional decoding algorithms cannot incorporate constraints during generation. Therefore, they can not ensure the output satisfies certain requirements that are important to certain applications. To this end, there exist two distinct lines of research focusing on controlled text generation which focus on constrained search algorithms and score-based sampling methods respectively. 

Constrained search algorithms enforce strict lexical constraints on the outputs by modifying the search space according to the constraints. Constrained beam search (CBS) algorithm~\citep{anderson-etal-2017-guided}, first proposed to track constraint satisfaction using a finite-state automaton.
%, encourages the model to fulfill unsatisfied constraints at each step. 
However, CBS requires the model to maintain a finite-state machine with $2^C$ states (where $C$ is the number of constraints), which results in significantly increased time complexity. 
To reduce the runtime overhead of CBS, \citet{hokamp-liu-2017-lexically} and ~\citet{post-vilar-2018-fast} introduced grid beam search (GBS) and dynamic beam allocation (DBA), respectively.
%GBS reduces inference to linear run-time complexity by grouping together hypotheses based on the number of constraints satisfied. 
%DBA further reduces the dependence on $C$ by fixing the beam size and splitting the beam into subgroups that correspond to different constraint satisfaction configurations.
\citet{lu-etal-2021-neurologic2} introduced NeuroLogic decoding, a constrained search algorithm that aims to satisfy rich logic-based lexical constraints expressed in conjunctive normal form and extend it to NeuroLogic A$^*$ decoding~\citep{lu-etal-2022-neurologic2} by incorporating lookahead heuristics. 
While existing constrained search algorithms significantly improve the constraint satisfaction rate, they still lead to slower generation speed and lower quality text because they prune the output distribution space aggressively and, thus, tend to collapse into globally suboptimal results. Moreover, most constrained search algorithms cannot incorporate constraint types other than lexical constraints.

Score-based sampling methods attempt to incorporate constraints by converting constraints into differentiable score functions. Specifically, score functions for soft constraints such as sentiment control can be implemented as the cross entropy loss of corresponding classifiers. Hard constraints such as lexical constraints can be modeled by a differentiable $n$-gram matching function~\citep{liu-etal-2022-dont}. 
%With score functions that encode different constraints, controlled text generation can be achieved by score-based generative models such as Langevin dynamics based models~\citep{song2022langevin,qin2022cold,kumar2022gradient} and diffusion models~\citep{diffusion,li2022diffusionlm} on either logit space or embedding space. 
Compared to constrained search algorithms, score-based sampling methods are more flexible because they can deal with more diverse constraint types and the composition of different constraints. However, score-based sampling methods do not come with constraint satisfaction guarantees and often lead to worse generation quality because the output distribution is modified~\citep{qin2022cold}. Moreover, score-based sampling methods lead to much slower generation speeds because multiple score-matching steps need to be performed. They also require carefully tuned relative weights between the scores of different constraints and the task-specific loss to achieve a good trade-off between the quality and the constraint satisfaction rate of output texts.

Apart from search-based and score-based methods, \citet{dinu-etal-2019-training} proposed a similar specialized training method that appends lexical constraints to the input when training the model. Another related work is the Similarly, CTRL~\citep{keskar2019ctrl} is pre-trained with structures that naturally co-occur with raw texts and can deal with certain kinds of constraints such as style and domain constraints.\looseness=-1

\section{Conclusion}

In this work, we introduce \method, a simple framework for controlled text generation which exploits the instruction following ability of large language models as an alternative to modifying the decoding procedure of the model. \method is faster and achieves better generation quality compared to state-of-the-art controlled generation methods while maintaining similar quality. 
\method also shows a strong few-shot generalization ability to unseen constraints, the ability to model the composition of multiple constraints, and can be seamlessly extended to conditional natural language generation tasks. 
%These results indicate prompt-based constrained generation is a promising alternative paradigm compared to existing decoding-time methods for controlled text generation.

%\section*{Limitations}

%\section*{Ethics Statement}

\section*{Acknowledgements}
Ryan Cotterell acknowledges support from the Swiss National Science Foundation (SNSF) as part of the ``The Forgotten Role of Inductive Bias in Interpretability'' project. EGW was supported by an ETH Postdoctoral Fellowship.

% Entries for the entire Anthology, followed by custom entries
\bibliography{anthology,custom}

\begin{thebibliography}{46}
\expandafter\ifx\csname natexlab\endcsname\relax\def\natexlab#1{#1}\fi

\bibitem[{Amini et~al.(2023)Amini, Du, and Cotterell}]{amini2023structured}
Afra Amini, Li~Du, and Ryan Cotterell. 2023.
\newblock \href {http://arxiv.org/abs/2306.03061} {Structured {V}oronoi
  sampling}.
\newblock \emph{arXiv preprint arXiv:2306.03061}.

\bibitem[{Anderson et~al.(2017)Anderson, Fernando, Johnson, and
  Gould}]{anderson-etal-2017-guided}
Peter Anderson, Basura Fernando, Mark Johnson, and Stephen Gould. 2017.
\newblock \href {https://doi.org/10.18653/v1/D17-1098} {Guided open vocabulary
  image captioning with constrained beam search}.
\newblock In \emph{Proceedings of the 2017 Conference on Empirical Methods in
  Natural Language Processing}, pages 936--945, Copenhagen, Denmark.
  Association for Computational Linguistics.

\bibitem[{Brown et~al.(2020)Brown, Mann, Ryder, Subbiah, Kaplan, Dhariwal,
  Neelakantan, Shyam, Sastry, Askell, Agarwal, Herbert-Voss, Krueger, Henighan,
  Child, Ramesh, Ziegler, Wu, Winter, Hesse, Chen, Sigler, Litwin, Gray, Chess,
  Clark, Berner, McCandlish, Radford, Sutskever, and Amodei}]{gpt3}
Tom Brown, Benjamin Mann, Nick Ryder, Melanie Subbiah, Jared Kaplan, Prafulla
  Dhariwal, Arvind Neelakantan, Pranav Shyam, Girish Sastry, Amanda Askell,
  Sandhini Agarwal, Ariel Herbert-Voss, Gretchen Krueger, Tom Henighan, Rewon
  Child, Aditya Ramesh, Daniel Ziegler, Jeffrey Wu, Clemens Winter, Chris
  Hesse, Mark Chen, Eric Sigler, Mateusz Litwin, Scott Gray, Benjamin Chess,
  Jack Clark, Christopher Berner, Sam McCandlish, Alec Radford, Ilya Sutskever,
  and Dario Amodei. 2020.
\newblock \href
  {https://proceedings.neurips.cc/paper/2020/file/1457c0d6bfcb4967418bfb8ac142f64a-Paper.pdf}
  {Language models are few-shot learners}.
\newblock In \emph{Advances in Neural Information Processing Systems},
  volume~33, pages 1877--1901.

\bibitem[{Dinu et~al.(2019)Dinu, Mathur, Federico, and
  Al-Onaizan}]{dinu-etal-2019-training}
Georgiana Dinu, Prashant Mathur, Marcello Federico, and Yaser Al-Onaizan. 2019.
\newblock \href {https://doi.org/10.18653/v1/P19-1294} {Training neural machine
  translation to apply terminology constraints}.
\newblock In \emph{Proceedings of the 57th Annual Meeting of the Association
  for Computational Linguistics}, pages 3063--3068, Florence, Italy.
  Association for Computational Linguistics.

\bibitem[{Du and Cardie(2018)}]{du-cardie-2018-harvesting}
Xinya Du and Claire Cardie. 2018.
\newblock \href {https://doi.org/10.18653/v1/P18-1177} {Harvesting
  paragraph-level question-answer pairs from {W}ikipedia}.
\newblock In \emph{Proceedings of the 56th Annual Meeting of the Association
  for Computational Linguistics (Volume 1: Long Papers)}, pages 1907--1917,
  Melbourne, Australia. Association for Computational Linguistics.

\bibitem[{Fan et~al.(2018)Fan, Lewis, and Dauphin}]{fan2018hierarchical}
Angela Fan, Mike Lewis, and Yann Dauphin. 2018.
\newblock \href {https://doi.org/10.18653/v1/P18-1082} {Hierarchical neural
  story generation}.
\newblock In \emph{Proceedings of the 56th Annual Meeting of the Association
  for Computational Linguistics (Volume 1: Long Papers)}, pages 889--898,
  Melbourne, Australia. Association for Computational Linguistics.

\bibitem[{Gan et~al.(2017)Gan, Gan, He, Gao, and Deng}]{gan2017stylenet}
Chuang Gan, Zhe Gan, Xiaodong He, Jianfeng Gao, and Li~Deng. 2017.
\newblock \href {https://ieeexplore.ieee.org/document/8099591} {{StyleNet}:
  {G}enerating attractive visual captions with styles}.
\newblock In \emph{Proceedings of the IEEE Conference on Computer Vision and
  Pattern Recognition}, pages 3137--3146.

\bibitem[{He and McAuley(2016)}]{he2016ups}
Ruining He and Julian McAuley. 2016.
\newblock \href {https://dl.acm.org/doi/10.1145/2872427.2883037} {Ups and
  downs: {M}odeling the visual evolution of fashion trends with one-class
  collaborative filtering}.
\newblock In \emph{Proceedings of the 25th International Conference on World
  Wide Web}, pages 507--517.

\bibitem[{Hochreiter and Schmidhuber(1997)}]{hochreiter1997long}
Sepp Hochreiter and J{\"u}rgen Schmidhuber. 1997.
\newblock \href
  {https://direct.mit.edu/neco/article-abstract/9/8/1735/6109/Long-Short-Term-Memory?redirectedFrom=fulltext}
  {Long short-term memory}.
\newblock \emph{Neural Computation}, 9(8):1735--1780.

\bibitem[{Hokamp and Liu(2017)}]{hokamp-liu-2017-lexically}
Chris Hokamp and Qun Liu. 2017.
\newblock \href {https://doi.org/10.18653/v1/P17-1141} {Lexically constrained
  decoding for sequence generation using grid beam search}.
\newblock In \emph{Proceedings of the 55th Annual Meeting of the Association
  for Computational Linguistics (Volume 1: Long Papers)}, pages 1535--1546,
  Vancouver, Canada. Association for Computational Linguistics.

\bibitem[{Holtzman et~al.(2020)Holtzman, Buys, Du, Forbes, and
  Choi}]{holtzman2019curious}
Ari Holtzman, Jan Buys, Li~Du, Maxwell Forbes, and Yejin Choi. 2020.
\newblock \href {https://openreview.net/pdf?id=rygGQyrFvH} {The curious case of
  neural text degeneration}.
\newblock In \emph{International Conference on Learning Representations}.

\bibitem[{Jin et~al.(2022)Jin, Jin, Hu, Vechtomova, and
  Mihalcea}]{jin-etal-2022-deep}
Di~Jin, Zhijing Jin, Zhiting Hu, Olga Vechtomova, and Rada Mihalcea. 2022.
\newblock \href {https://doi.org/10.1162/coli_a_00426} {Deep learning for text
  style transfer: A survey}.
\newblock \emph{Computational Linguistics}, 48(1):155--205.

\bibitem[{Keskar et~al.(2019)Keskar, McCann, Varshney, Xiong, and
  Socher}]{keskar2019ctrl}
Nitish~Shirish Keskar, Bryan McCann, Lav~R. Varshney, Caiming Xiong, and
  Richard Socher. 2019.
\newblock \href {https://arxiv.org/abs/1909.05858} {{CTRL}: {A} conditional
  transformer language model for controllable generation}.
\newblock \emph{arXiv preprint arXiv:1909.05858}.

\bibitem[{Kingma and Ba(2015)}]{kingma2014adam}
Diederik~P. Kingma and Jimmy Ba. 2015.
\newblock \href {http://dblp.uni-trier.de/db/conf/iclr/iclr2015.html#KingmaB14}
  {Adam: A method for stochastic optimization}.
\newblock In \emph{International Conference on Learning Representations}.

\bibitem[{Kitaev and Klein(2018)}]{kitaev-klein-2018-constituency}
Nikita Kitaev and Dan Klein. 2018.
\newblock \href {https://doi.org/10.18653/v1/P18-1249} {Constituency parsing
  with a self-attentive encoder}.
\newblock In \emph{Proceedings of the 56th Annual Meeting of the Association
  for Computational Linguistics (Volume 1: Long Papers)}, pages 2676--2686,
  Melbourne, Australia. Association for Computational Linguistics.

\bibitem[{Koehn et~al.(2007)Koehn, Hoang, Birch, Callison-Burch, Federico,
  Bertoldi, Cowan, Shen, Moran, Zens, Dyer, Bojar, Constantin, and
  Herbst}]{koehn-etal-2007-moses}
Philipp Koehn, Hieu Hoang, Alexandra Birch, Chris Callison-Burch, Marcello
  Federico, Nicola Bertoldi, Brooke Cowan, Wade Shen, Christine Moran, Richard
  Zens, Chris Dyer, Ond{\v{r}}ej Bojar, Alexandra Constantin, and Evan Herbst.
  2007.
\newblock \href {https://aclanthology.org/P07-2045} {{M}oses: Open source
  toolkit for statistical machine translation}.
\newblock In \emph{Proceedings of the 45th Annual Meeting of the Association
  for Computational Linguistics Companion Volume Proceedings of the Demo and
  Poster Sessions}, pages 177--180, Prague, Czech Republic. Association for
  Computational Linguistics.

\bibitem[{Kumar et~al.(2022)Kumar, Paria, and Tsvetkov}]{kumar2022gradient}
Sachin Kumar, Biswajit Paria, and Yulia Tsvetkov. 2022.
\newblock \href {https://aclanthology.org/2022.emnlp-main.144} {Gradient-based
  constrained sampling from language models}.
\newblock In \emph{Proceedings of the 2022 Conference on Empirical Methods in
  Natural Language Processing}, pages 2251--2277, Abu Dhabi, United Arab
  Emirates. Association for Computational Linguistics.

\bibitem[{Li et~al.(2018)Li, Jia, He, and Liang}]{li-etal-2018-delete}
Juncen Li, Robin Jia, He~He, and Percy Liang. 2018.
\newblock \href {https://doi.org/10.18653/v1/N18-1169} {Delete, retrieve,
  generate: a simple approach to sentiment and style transfer}.
\newblock In \emph{Proceedings of the 2018 Conference of the North {A}merican
  Chapter of the Association for Computational Linguistics: Human Language
  Technologies, Volume 1 (Long Papers)}, pages 1865--1874, New Orleans,
  Louisiana. Association for Computational Linguistics.

\bibitem[{Li et~al.(2022)Li, Thickstun, Gulrajani, Liang, and
  Hashimoto}]{li2022diffusionlm}
Xiang~Lisa Li, John Thickstun, Ishaan Gulrajani, Percy Liang, and Tatsunori
  Hashimoto. 2022.
\newblock \href {https://openreview.net/forum?id=3s9IrEsjLyk} {Diffusion-{LM}
  improves controllable text generation}.
\newblock In \emph{Advances in Neural Information Processing Systems}.

\bibitem[{Lin et~al.(2020)Lin, Zhou, Shen, Zhou, Bhagavatula, Choi, and
  Ren}]{lin-etal-2020-commongen}
Bill~Yuchen Lin, Wangchunshu Zhou, Ming Shen, Pei Zhou, Chandra Bhagavatula,
  Yejin Choi, and Xiang Ren. 2020.
\newblock \href {https://doi.org/10.18653/v1/2020.findings-emnlp.165}
  {{C}ommon{G}en: A constrained text generation challenge for generative
  commonsense reasoning}.
\newblock In \emph{Findings of the Association for Computational Linguistics:
  EMNLP 2020}, pages 1823--1840, Online. Association for Computational
  Linguistics.

\bibitem[{Liu et~al.(2022)Liu, Yang, Tao, Liang, Bao, Li, He, Cui, and
  Hu}]{liu-etal-2022-dont}
Guangyi Liu, Zichao Yang, Tianhua Tao, Xiaodan Liang, Junwei Bao, Zhen Li,
  Xiaodong He, Shuguang Cui, and Zhiting Hu. 2022.
\newblock \href {https://doi.org/10.18653/v1/2022.naacl-main.150} {Don{'}t take
  it literally: An edit-invariant sequence loss for text generation}.
\newblock In \emph{Proceedings of the 2022 Conference of the North American
  Chapter of the Association for Computational Linguistics: Human Language
  Technologies}, pages 2055--2078, Seattle, United States. Association for
  Computational Linguistics.

\bibitem[{Liu et~al.(2019)Liu, Ott, Goyal, Du, Joshi, Chen, Levy, Lewis,
  Zettlemoyer, and Stoyanov}]{liu2019roberta}
Yinhan Liu, Myle Ott, Naman Goyal, Jingfei Du, Mandar Joshi, Danqi Chen, Omer
  Levy, Mike Lewis, Luke Zettlemoyer, and Veselin Stoyanov. 2019.
\newblock \href {https://arxiv.org/abs/1907.11692} {{RoBERTa}: {A} robustly
  optimized {BERT} pretraining approach}.
\newblock \emph{arXiv preprint arXiv:1907.11692}.

\bibitem[{Lu et~al.(2022)Lu, Welleck, West, Jiang, Kasai, Khashabi, Le~Bras,
  Qin, Yu, Zellers, Smith, and Choi}]{lu-etal-2022-neurologic2}
Ximing Lu, Sean Welleck, Peter West, Liwei Jiang, Jungo Kasai, Daniel Khashabi,
  Ronan Le~Bras, Lianhui Qin, Youngjae Yu, Rowan Zellers, Noah~A. Smith, and
  Yejin Choi. 2022.
\newblock \href {https://doi.org/10.18653/v1/2022.naacl-main.57}
  {{N}euro{L}ogic {A$^*$esque} decoding: {C}onstrained text generation with
  lookahead heuristics}.
\newblock In \emph{Proceedings of the 2022 Conference of the North American
  Chapter of the Association for Computational Linguistics: Human Language
  Technologies}, pages 780--799, Seattle, United States. Association for
  Computational Linguistics.

\bibitem[{Lu et~al.(2021)Lu, West, Zellers, Le~Bras, Bhagavatula, and
  Choi}]{lu-etal-2021-neurologic2}
Ximing Lu, Peter West, Rowan Zellers, Ronan Le~Bras, Chandra Bhagavatula, and
  Yejin Choi. 2021.
\newblock \href {https://doi.org/10.18653/v1/2021.naacl-main.339}
  {{N}euro{L}ogic decoding: {(Un)supervised} neural text generation with
  predicate logic constraints}.
\newblock In \emph{Proceedings of the 2021 Conference of the North American
  Chapter of the Association for Computational Linguistics: Human Language
  Technologies}, pages 4288--4299, Online. Association for Computational
  Linguistics.

\bibitem[{Madaan et~al.(2020)Madaan, Setlur, Parekh, Poczos, Neubig, Yang,
  Salakhutdinov, Black, and Prabhumoye}]{madaan-etal-2020-politeness}
Aman Madaan, Amrith Setlur, Tanmay Parekh, Barnabas Poczos, Graham Neubig,
  Yiming Yang, Ruslan Salakhutdinov, Alan~W Black, and Shrimai Prabhumoye.
  2020.
\newblock \href {https://doi.org/10.18653/v1/2020.acl-main.169} {Politeness
  transfer: A tag and generate approach}.
\newblock In \emph{Proceedings of the 58th Annual Meeting of the Association
  for Computational Linguistics}, pages 1869--1881, Online. Association for
  Computational Linguistics.

\bibitem[{Meister et~al.(2023)Meister, Pimentel, Wiher, and
  Cotterell}]{meister2022locally}
Clara Meister, Tiago Pimentel, Gian Wiher, and Ryan Cotterell. 2023.
\newblock \href {https://doi.org/10.1162/tacl_a_00536} {{Locally Typical
  Sampling}}.
\newblock \emph{Transactions of the Association for Computational Linguistics},
  11:102--121.

\bibitem[{Min et~al.(2022)Min, Lewis, Zettlemoyer, and
  Hajishirzi}]{min-etal-2022-metaicl}
Sewon Min, Mike Lewis, Luke Zettlemoyer, and Hannaneh Hajishirzi. 2022.
\newblock \href {https://doi.org/10.18653/v1/2022.naacl-main.201} {{M}eta{ICL}:
  Learning to learn in context}.
\newblock In \emph{Proceedings of the 2022 Conference of the North American
  Chapter of the Association for Computational Linguistics: Human Language
  Technologies}, pages 2791--2809, Seattle, United States. Association for
  Computational Linguistics.

\bibitem[{Post and Vilar(2018)}]{post-vilar-2018-fast}
Matt Post and David Vilar. 2018.
\newblock \href {https://doi.org/10.18653/v1/N18-1119} {Fast lexically
  constrained decoding with dynamic beam allocation for neural machine
  translation}.
\newblock In \emph{Proceedings of the 2018 Conference of the North {A}merican
  Chapter of the Association for Computational Linguistics: Human Language
  Technologies, Volume 1 (Long Papers)}, pages 1314--1324, New Orleans,
  Louisiana. Association for Computational Linguistics.

\bibitem[{Pryzant et~al.(2020)Pryzant, Martinez, Dass, Kurohashi, Jurafsky, and
  Yang}]{pryzant2020automatically}
Reid Pryzant, Richard~Diehl Martinez, Nathan Dass, Sadao Kurohashi, Dan
  Jurafsky, and Diyi Yang. 2020.
\newblock \href {https://ojs.aaai.org/index.php/AAAI/article/view/5385/5241}
  {Automatically neutralizing subjective bias in text}.
\newblock In \emph{Proceedings of the AAAI Conference on Artificial
  Intelligence}, volume~34, pages 480--489.

\bibitem[{Qin et~al.(2022)Qin, Welleck, Khashabi, and Choi}]{qin2022cold}
Lianhui Qin, Sean Welleck, Daniel Khashabi, and Yejin Choi. 2022.
\newblock \href {https://openreview.net/forum?id=TiZYrQ-mPup} {{COLD} decoding:
  {Energy-based} constrained text generation with {L}angevin dynamics}.
\newblock In \emph{Advances in Neural Information Processing Systems}.

\bibitem[{Radford et~al.(2018)Radford, Narasimhan, Salimans, Sutskever
  et~al.}]{gpt}
Alec Radford, Karthik Narasimhan, Tim Salimans, Ilya Sutskever, et~al. 2018.
\newblock \href {https://openai.com/research/language-unsupervised} {Improving
  language understanding by generative pre-training}.

\bibitem[{Radford et~al.(2019)Radford, Wu, Child, Luan, Amodei, Sutskever
  et~al.}]{gpt2}
Alec Radford, Jeffrey Wu, Rewon Child, David Luan, Dario Amodei, Ilya
  Sutskever, et~al. 2019.
\newblock \href
  {https://d4mucfpksywv.cloudfront.net/better-language-models/language_models_are_unsupervised_multitask_learners.pdf}
  {Language models are unsupervised multitask learners}.
\newblock \emph{OpenAI Blog}, 1(8):9.

\bibitem[{Raffel et~al.(2020)Raffel, Shazeer, Roberts, Lee, Narang, Matena,
  Zhou, Li, and Liu}]{raffel2020exploring}
Colin Raffel, Noam Shazeer, Adam Roberts, Katherine Lee, Sharan Narang, Michael
  Matena, Yanqi Zhou, Wei Li, and Peter~J. Liu. 2020.
\newblock \href {http://jmlr.org/papers/v21/20-074.html} {Exploring the limits
  of transfer learning with a unified text-to-text transformer}.
\newblock \emph{Journal of Machine Learning Research}, 21(140):1--67.

\bibitem[{Rajpurkar et~al.(2016)Rajpurkar, Zhang, Lopyrev, and
  Liang}]{rajpurkar-etal-2016-squad}
Pranav Rajpurkar, Jian Zhang, Konstantin Lopyrev, and Percy Liang. 2016.
\newblock \href {https://doi.org/10.18653/v1/D16-1264} {{SQ}u{AD}: 100,000+
  questions for machine comprehension of text}.
\newblock In \emph{Proceedings of the 2016 Conference on Empirical Methods in
  Natural Language Processing}, pages 2383--2392, Austin, Texas. Association
  for Computational Linguistics.

\bibitem[{Rao and Tetreault(2018)}]{rao-tetreault-2018-dear}
Sudha Rao and Joel Tetreault. 2018.
\newblock \href {https://doi.org/10.18653/v1/N18-1012} {Dear sir or madam, may
  {I} introduce the {GYAFC} dataset: Corpus, benchmarks and metrics for
  formality style transfer}.
\newblock In \emph{Proceedings of the 2018 Conference of the North {A}merican
  Chapter of the Association for Computational Linguistics: Human Language
  Technologies, Volume 1 (Long Papers)}, pages 129--140, New Orleans,
  Louisiana. Association for Computational Linguistics.

\bibitem[{Reid et~al.(2022)Reid, Zhong, Gururangan, and
  Zettlemoyer}]{reid2022m2d2}
Machel Reid, Victor Zhong, Suchin Gururangan, and Luke Zettlemoyer. 2022.
\newblock \href {https://aclanthology.org/2022.emnlp-main.63} {{M}2{D}2: A
  massively multi-domain language modeling dataset}.
\newblock In \emph{Proceedings of the 2022 Conference on Empirical Methods in
  Natural Language Processing}, pages 964--975, Abu Dhabi, United Arab
  Emirates. Association for Computational Linguistics.

\bibitem[{Sanh et~al.(2022)Sanh, Webson, Raffel, Bach, Sutawika, Alyafeai,
  Chaffin, Stiegler, Raja, Dey, Bari, Xu, Thakker, Sharma, Szczechla, Kim,
  Chhablani, Nayak, Datta, Chang, Jiang, Wang, Manica, Shen, Yong, Pandey,
  Bawden, Wang, Neeraj, Rozen, Sharma, Santilli, Fevry, Fries, Teehan, Scao,
  Biderman, Gao, Wolf, and Rush}]{sanh2022multitask}
Victor Sanh, Albert Webson, Colin Raffel, Stephen Bach, Lintang Sutawika, Zaid
  Alyafeai, Antoine Chaffin, Arnaud Stiegler, Arun Raja, Manan Dey, M~Saiful
  Bari, Canwen Xu, Urmish Thakker, Shanya~Sharma Sharma, Eliza Szczechla,
  Taewoon Kim, Gunjan Chhablani, Nihal Nayak, Debajyoti Datta, Jonathan Chang,
  Mike Tian-Jian Jiang, Han Wang, Matteo Manica, Sheng Shen, Zheng~Xin Yong,
  Harshit Pandey, Rachel Bawden, Thomas Wang, Trishala Neeraj, Jos Rozen,
  Abheesht Sharma, Andrea Santilli, Thibault Fevry, Jason~Alan Fries, Ryan
  Teehan, Teven~Le Scao, Stella Biderman, Leo Gao, Thomas Wolf, and Alexander~M
  Rush. 2022.
\newblock \href {https://openreview.net/forum?id=9Vrb9D0WI4} {Multitask
  prompted training enables zero-shot task generalization}.
\newblock In \emph{International Conference on Learning Representations}.

\bibitem[{Tran et~al.(2020)Tran, Zhang, and Soleymani}]{tran-etal-2020-towards}
Minh Tran, Yipeng Zhang, and Mohammad Soleymani. 2020.
\newblock \href {https://doi.org/10.18653/v1/2020.coling-main.190} {Towards a
  friendly online community: An unsupervised style transfer framework for
  profanity redaction}.
\newblock In \emph{Proceedings of the 28th International Conference on
  Computational Linguistics}, pages 2107--2114, Barcelona, Spain (Online).
  International Committee on Computational Linguistics.

\bibitem[{Vaswani et~al.(2017)Vaswani, Shazeer, Parmar, Uszkoreit, Jones,
  Gomez, Kaiser, and Polosukhin}]{transformer}
Ashish Vaswani, Noam Shazeer, Niki Parmar, Jakob Uszkoreit, Llion Jones,
  Aidan~N. Gomez, {\L}ukasz Kaiser, and Illia Polosukhin. 2017.
\newblock \href
  {https://proceedings.neurips.cc/paper/2017/file/3f5ee243547dee91fbd053c1c4a845aa-Paper.pdf}
  {Attention is all you need}.
\newblock In \emph{Advances in Neural Information Processing Systems},
  volume~30.

\bibitem[{Voigt et~al.(2018)Voigt, Jurgens, Prabhakaran, Jurafsky, and
  Tsvetkov}]{voigt2018rtgender}
Rob Voigt, David Jurgens, Vinodkumar Prabhakaran, Dan Jurafsky, and Yulia
  Tsvetkov. 2018.
\newblock \href {https://aclanthology.org/L18-1445/} {{R}t{G}ender: A corpus
  for studying differential responses to gender}.
\newblock In \emph{Proceedings of the Eleventh International Conference on
  Language Resources and Evaluation ({LREC} 2018)}, Miyazaki, Japan. European
  Language Resources Association (ELRA).

\bibitem[{Wei et~al.(2022)Wei, Bosma, Zhao, Guu, Yu, Lester, Du, Dai, and
  Le}]{wei2022finetuned}
Jason Wei, Maarten Bosma, Vincent Zhao, Kelvin Guu, Adams~Wei Yu, Brian Lester,
  Nan Du, Andrew~M. Dai, and Quoc~V. Le. 2022.
\newblock \href {https://openreview.net/forum?id=gEZrGCozdqR} {Finetuned
  language models are zero-shot learners}.
\newblock In \emph{International Conference on Learning Representations}.

\bibitem[{Xu et~al.(2018)Xu, Sun, Zeng, Zhang, Ren, Wang, and
  Li}]{xu-etal-2018-unpaired}
Jingjing Xu, Xu~Sun, Qi~Zeng, Xiaodong Zhang, Xuancheng Ren, Houfeng Wang, and
  Wenjie Li. 2018.
\newblock \href {https://doi.org/10.18653/v1/P18-1090} {Unpaired
  sentiment-to-sentiment translation: A cycled reinforcement learning
  approach}.
\newblock In \emph{Proceedings of the 56th Annual Meeting of the Association
  for Computational Linguistics (Volume 1: Long Papers)}, pages 979--988,
  Melbourne, Australia. Association for Computational Linguistics.

\bibitem[{Zhang et~al.(2022)Zhang, Roller, Goyal, Artetxe, Chen, Chen, Dewan,
  Diab, Li, Lin et~al.}]{zhang2022opt}
Susan Zhang, Stephen Roller, Naman Goyal, Mikel Artetxe, Moya Chen, Shuohui
  Chen, Christopher Dewan, Mona Diab, Xian Li, Xi~Victoria Lin, et~al. 2022.
\newblock \href {https://arxiv.org/abs/2205.01068} {{OPT}: {Open} pre-trained
  transformer language models}.
\newblock \emph{arXiv preprint arXiv:2205.01068}.

\bibitem[{Zhou et~al.(2019)Zhou, Ge, Xu, Wei, and Zhou}]{zhou-etal-2019-bert}
Wangchunshu Zhou, Tao Ge, Ke~Xu, Furu Wei, and Ming Zhou. 2019.
\newblock \href {https://doi.org/10.18653/v1/P19-1328} {{BERT}-based lexical
  substitution}.
\newblock In \emph{Proceedings of the 57th Annual Meeting of the Association
  for Computational Linguistics}, pages 3368--3373, Florence, Italy.
  Association for Computational Linguistics.

\bibitem[{Zhou et~al.(2021)Zhou, Lee, Selvam, Lee, and
  Ren}]{zhou2021pretraining}
Wangchunshu Zhou, Dong-Ho Lee, Ravi~Kiran Selvam, Seyeon Lee, and Xiang Ren.
  2021.
\newblock \href {https://openreview.net/forum?id=3k20LAiHYL2} {Pre-training
  text-to-text transformers for concept-centric common sense}.
\newblock In \emph{International Conference on Learning Representations}.

\bibitem[{Zhu et~al.(2010)Zhu, Bernhard, and
  Gurevych}]{zhu-etal-2010-monolingual2}
Zhemin Zhu, Delphine Bernhard, and Iryna Gurevych. 2010.
\newblock \href {https://aclanthology.org/C10-1152} {A monolingual tree-based
  translation model for sentence simplification}.
\newblock In \emph{Proceedings of the 23rd International Conference on
  Computational Linguistics}, pages 1353--1361, Beijing, China.

\end{thebibliography}
\bibliographystyle{acl_natbib}

\newpage
\appendix
\onecolumn
\section{Appendix}
\label{sec:appendix}

\begin{table*}[!h]
\centering
\begin{adjustbox}{width=0.95\textwidth}
\begin{tabular}{ccp{440pt}}
\toprule[2pt]
% \multicolumn{2}{c}{\textsc{lexical}} \\
% \midrule[1pt]
\task{lexical} & \cellcolor{gray!20} \textsc{constraints}  & \cellcolor{gray!20} I $\wedge$ time  $\wedge$ television $\wedge$ game  \\
& \textsc{NeuroLogic} & I watch television and play game for a time.   \\
& \textsc{Diffusion-LM} & I play game and television at this time.  \\
& \textsc{\method} & I enjoy watching television and playing video games in my free time. \\
\midrule[2pt]
% \multicolumn{2}{c}{\textsc{syntax}} \\
% \midrule[1pt]
\task{syntax} & \cellcolor{gray!20} \textsc{constraints}  & \cellcolor{gray!20} \textsc{verb} \textsc{det} \textsc{adj} \textsc{noun} \textsc{prep} \textsc{noun} \textsc{noun} \textsc{noun} \\
& \textsc{COLD} & Look this white bear in the forest stone  \\
& \textsc{Diffusion-LM} & Bring the new computer to storage room door  \\
& \textsc{\method} & Read this new article about Aircraft System design. \\
\midrule[2pt]
% \multicolumn{2}{c}{\textsc{semantic}} \\
% \midrule[1pt]
\task{semantic} & \cellcolor{gray!20} \textsc{constraints}  & \cellcolor{gray!20}  the sentiment being negative \\
& \textsc{COLD} & This restaurant is not good. It would just soon be lacking of quality for the meat and sausage.    \\
& \textsc{Diffusion-LM} & The service is very disappointing and the quality of the food like meat and wine is also not of good level.  \\
& \textsc{\method} & This article is really boring. It is completely non sense to me and my collegues.  \\
\midrule[2pt]
% \multicolumn{2}{c}{\textsc{sytle}} \\
% \midrule[1pt]
\task{style} & \cellcolor{gray!20} \textsc{constraints}  & \cellcolor{gray!20} the style being polite\\
& \textsc{COLD} & You are honored to come there for the meeting \\
& \textsc{Diffusion-LM} & I would be pleasure to receive my money back sincerely \\
& \textsc{Diffusion-LM} & Could you please help me check the status of my order \\
%\midrule[2pt]
% \multicolumn{2}{c}{\textsc{length}} \\
% \midrule[1pt]
\bottomrule[2pt]
\end{tabular}
\end{adjustbox}
    \caption{Samples generated with \method and compared models for different control tasks.}
    \label{tab:samples}
    \vspace{-5pt}
\end{table*}
\subsection{Limitations and Potential Social Impacts}
One limitation is that we only consider a few types of constraints in our experiments and only test \method with T0 as the backbone model. 
It would be helpful to test \method on more diverse constraint types and backbone models of different sizes to get an even better picture of its utility.
As for potential negative social impacts, \method could be used by malicious users by writing instructions that encourage the model to generate toxic or biased texts. 
This is a common potential risk of controlled text generation methods. 
%We believe this risk could be partially reduced by applying bias m or unlearning methods to make the model unable to generate harmful texts.

\subsection{Implementation Details}

\paragraph{Training Data}
As described in Section \ref{sec:data} and \ref{sec:composition}, we train the fine-tune on 1 million examples containing a single constraint and 1 million examples containing multiple constraints (ranging from 2 to 5).
When fine-tuning for conditional natural language generation tasks, we use all the training examples in the original datasets and synthesize training data with the lexical constraints using the method described in Section \ref{sec:data}. We fine-tune the model with a learning rate of 5e-5 with other hyperparameters unchanged compared to that described in Section \ref{sec:training}.

\paragraph{Baseline Training}
We train the CFT baseline with the exact hyperparameters used during the training of \method, which is empirically found to work well for both conditional fine-tuning and our approach. 
For other baselines, we start from the hyperparameters provided in the corresponding papers and do a grid search over other possible values on the development set.\looseness=-1

\paragraph{Constraint Composition Evaluation}
In Section \ref{sec:compose_eval}, we compose constraints from different categories to avoid potential conflicts. Specifically, we first select 1000 test examples that have at least one valid constraint in each category. We then test the model by randomly sampling and composing 2,3,4,5 constraints.

\subsection{Examples of Generated Text}
We also present some samples of \method and compared baselines in Table \ref{tab:samples} for qualitative analysis. 
%We find that while all compared models mostly succeed the constraint, \method generate much more fluent and interesting samples. The performance is measured by first normalizing \success \ and \fluency \ with the averaged single constraint performance and then calculating the average. 
%For score-based baselines, the relative weights are tuned with random search.
\end{document}